\documentclass[sigconf,review=false, timestamp=false, nonacm, screen]{acmart}

\AtBeginDocument{%
  \providecommand\BibTeX{{%
    \normalfont B\kern-0.5em{\scshape i\kern-0.25em b}\kern-0.8em\TeX}}}

\setcopyright{acmlicensed}
\copyrightyear{2018}
\acmYear{2018}
\acmDOI{XXXXXXX.XXXXXXX}

\newcommand{\DatasetName}{Klarna Product Page Dataset}
\newcommand{\TrainingName}{CNTP}

\usepackage{microtype}
\usepackage{graphicx}
\usepackage{subfigure}
\usepackage{xcolor}
\usepackage{dirtree}
\usepackage{natbib}

\usepackage{pgfplots}
\usepackage{pgfplotstable}
\usepackage{tableaucolors}
\pgfplotsset{compat=1.17}

\usetikzlibrary{shapes,shapes.geometric,arrows,fit,calc,automata,backgrounds, patterns}

\tikzset{
        /tikz/every even column/.append style={column sep=1em},
    }
\pgfplotsset{
    every axis/every ticks/.append style={fontsize=\small},
    every axis legend/.append style={
            font=\sc\small,
            {/tikz/every even column/.append style={column sep=1em}},
        }
}
\tikzstyle{sc0}=[very thick,densely dashed, C6]
\tikzstyle{sc1}=[very thick,densely dashed, C7]
\tikzstyle{sc2}=[very thick,densely dashed, C8]
\tikzstyle{sc3}=[very thick,densely dashed, C9]
\tikzstyle{sc4}=[very thick,densely dashed, C0]
\tikzstyle{sc5}=[very thick,densely dashed, C1]
\tikzstyle{sc6}=[very thick, C6]
\tikzstyle{sc7}=[very thick, C7]
\tikzstyle{sc8}=[very thick, C8]
\tikzstyle{sc9}=[very thick, C9]
\tikzstyle{sc10}=[very thick, C0]
\tikzstyle{sc11}=[very thick, C1]

\usetikzlibrary{positioning}

\usepackage[stable]{footmisc}

\usepackage[utf8]{inputenc} 
\usepackage[T1]{fontenc}    
\usepackage{hyperref}       
\usepackage{url}            
\usepackage{amsfonts}       
\usepackage{nicefrac}       

\usepackage{xcolor}         
\usepackage{svg}

\newcommand{\bx}{\mathbf{x}}
\newcommand{\bz}{\mathbf{z}}
\newcommand{\bb}{\mathbf{b}}

\newcommand{\bc}{\mathbf{c}}
\newcommand{\bh}{\mathbf{h}}

\newcommand{\bk}{\mathbf{k}}
\newcommand{\by}{\mathbf{y}}

\newcommand{\bw}{\mathbf{w}}
\newcommand{\bV}{\mathbf{V}}
\newcommand{\bH}{\mathbf{H}}

\newcommand{\bW}{\mathbf{W}}

\tikzstyle{s0}=[very thick, C6]
\tikzstyle{s1}=[very thick, C7]
\tikzstyle{s2}=[very thick, C8]
\tikzstyle{s3}=[very thick, C9]
\tikzstyle{s4}=[very thick, C0]
\tikzstyle{s5}=[very thick, C1]
\tikzstyle{s6}=[very thick, mark=*, C6]
\tikzstyle{s7}=[very thick, mark=*, C7]
\tikzstyle{s8}=[very thick, mark=*, C8]
\tikzstyle{s9}=[very thick, mark=*, C9]
\tikzstyle{s10}=[very thick, mark=*, C0]
\tikzstyle{s11}=[very thick, mark=*, C1]

\tikzstyle{sc0}=[very thick,densely dashed, C6]
\tikzstyle{sc1}=[very thick,densely dashed, C7]
\tikzstyle{sc2}=[very thick,densely dashed, C8]
\tikzstyle{sc3}=[very thick,densely dashed, C9]
\tikzstyle{sc4}=[very thick,densely dashed, C0]
\tikzstyle{sc5}=[very thick,densely dashed, C1]
\tikzstyle{sc6}=[very thick, C6]
\tikzstyle{sc7}=[very thick, C7]
\tikzstyle{sc8}=[very thick, C8]
\tikzstyle{sc9}=[very thick, C9]
\tikzstyle{sc10}=[very thick, C0]
\tikzstyle{sc11}=[very thick, C1]

\newlength{\mybarwidth}

\usepackage{amsmath}
\usepackage{mathtools}
\usepackage{amsthm}

\usepackage{lipsum}
\usepackage{float}
\usepackage{adjustbox} 
\usepackage{algorithm}
\usepackage{algpseudocode} 
\usepackage{multicol}
\usepackage{commath}
\usepackage{wrapfig}
\usepackage{comment}
\usepackage{caption}
\usepackage{dblfloatfix}
\usepackage{booktabs}    
\usepackage{graphicx}

\usepackage{hyperref}
\hypersetup{
    colorlinks=true,
    linkcolor=blue,
    filecolor=magenta,      
    urlcolor=cyan,
}

\begin{document}

\title{The Klarna Product Page Dataset: Web Element Nomination with Graph Neural Networks and Large Language Models }

\author{Alexandra~Hotti}
\affiliation{%
  \institution{KTH Royal Institute of Technology}
  \institution{Klarna}
  \city{Stockholm}
  \country{Sweden}
}
\email{hotti@kth.se}

\author{Riccardo~S.~Risuleo}
\affiliation{%
 \institution{Klarna}
  \city{Stockholm}
  \country{Sweden}
}
\email{riccardo.risuleo@klarna.com}

\author{Stefan~Magureanu}
\affiliation{%
 \institution{Klarna}
  \city{Stockholm}
  \country{Sweden}
}
\email{stefan.magureanu@klarna.com}

\author{Aref~Moradi}
\affiliation{%
 \institution{Klarna}
  \city{Stockholm}
  \country{Sweden}
}
\email{aref.moradi@klarna.com}

\author{Jens~Lagergren}
\affiliation{%
  \institution{SciLifeLab, School of EECS}
  \institution{KTH Royal Institute of Technology}
  \city{Stockholm}
  \country{Sweden}
}
\email{jens.lagergren@scilifelab.se}

\renewcommand{\shortauthors}{Hotti, et al.}


\begin{abstract}
Web automation holds the potential to revolutionize how users interact with the digital world, offering unparalleled assistance and simplifying tasks via sophisticated computational methods. Central to this evolution is the web element nomination task, which entails identifying unique elements on webpages. Unfortunately, the development of algorithmic designs for web automation is hampered by the scarcity of comprehensive and realistic datasets that reflect the complexity faced by real-world applications on the Web. To address this, we introduce the Klarna Product Page Dataset, a comprehensive and diverse collection of webpages that surpasses existing datasets in richness and variety. The dataset features 51,701 manually labeled product pages from 8,175 e-commerce websites across eight geographic regions, accompanied by a dataset of rendered page screenshots. To initiate research on the Klarna Product Page Dataset, we empirically benchmark a range of Graph Neural Networks (GNNs) on the web element nomination task. We make three important contributions. First, we found that a simple Convolutional GNN (GCN) outperforms complex state-of-the-art nomination methods. Second, we introduce a training refinement procedure that involves identifying a small number of relevant elements from each page using the aforementioned GCN. These elements are then passed to a large language model for the final nomination. This procedure significantly improves the nomination accuracy by 16.8 percentage points on our challenging dataset, without any need for fine-tuning. Finally, in response to another prevalent challenge in this field – the abundance of training methodologies suitable for element nomination – we introduce the \emph{Challenge Nomination Training Procedure}, a novel training approach that further boosts nomination accuracy.
\end{abstract}

\maketitle

\section{Introduction}\label{sec:intro}

In the rapidly advancing field of web automation \citep{yao2022webshop, Lin_2020, webformer, deng2023mind2web}, the overarching goal is to provide users with seamless assistance when completing mundane tasks online. By harnessing the full potential of web automation, users will no longer be required to manually navigate through web workflows. Instead, intrusive pop-ups will be preemptively closed, and all forms will be automatically filled in, irrespective of the specific website being accessed. One factor driving the recent surge in interest in web automation is the rise of large language models (LLMs). The impressive performance demonstrated by LLMs, combined with the fact that models such as GPT-4 \cite{openai2023gpt4} have been trained on HTML data, has catalyzed a surge in research focused on using LLMs as, for instance, foundation models on webpage datasets \cite{furuta2023multimodal, zhao2023expel, deng2023mind2web, zheng2024gpt, wang2023survey, kim2023language, zheng, gur2023understanding}.

However, a significant barrier to progress in this field is the scarcity of publicly accessible, large-scale, and realistic webpage datasets. Previous works have been restricted to simplified, simulated webpages \cite{gur2023understanding, yao2023webshop, pmlr-v70-shi17a}, or limited to pages from a narrow selection of websites \cite{swde, yao2023webshop}. These datasets fail to capture the challenges faced by real applications on the Web, which consists of tens of millions of websites, each featuring thousands of elements and layout patterns that can vary significantly across different websites. An explanation for the scarcity of datasets might be that collecting and annotating this type of dataset is a costly and time-consuming undertaking, demanding a set of skills more commonly possessed by web developers than by machine learning researchers.

\begin{figure*}[t]
    \centering
    \begin{subfigure}{}
    \includegraphics[width=0.18\textwidth, trim= 0 38cm 0 0, clip]{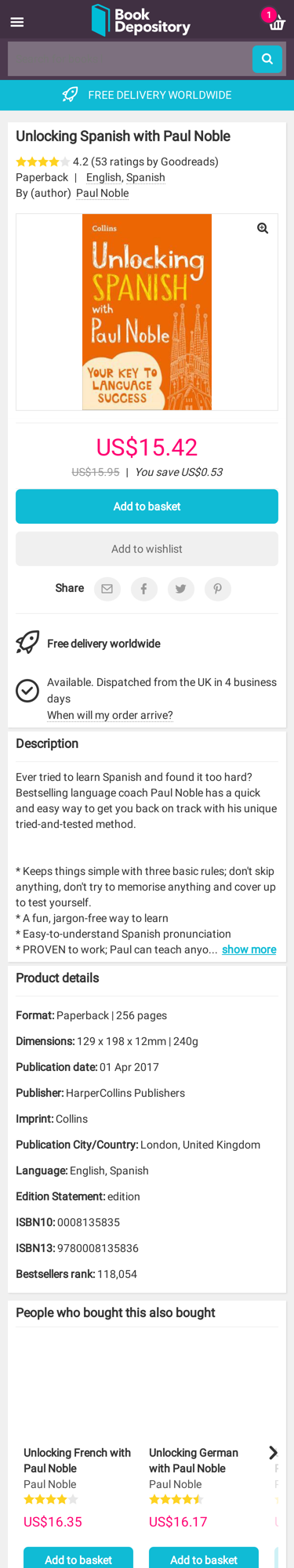}
        \label{fig:subfig1}
    \end{subfigure}
    \hfill
    \begin{subfigure}{}
    \includegraphics[width=0.18\textwidth, trim= 0 38cm 0 0, clip]{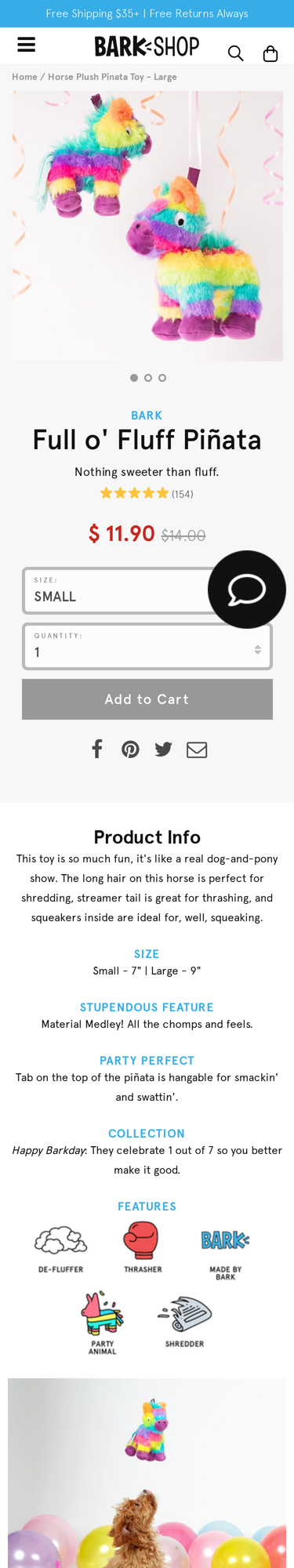}
        \label{fig:subfig2}
    \end{subfigure}
    \hfill
    \begin{subfigure}{}
    \includegraphics[width=0.18\textwidth, trim= 0 38cm 0 0, clip]{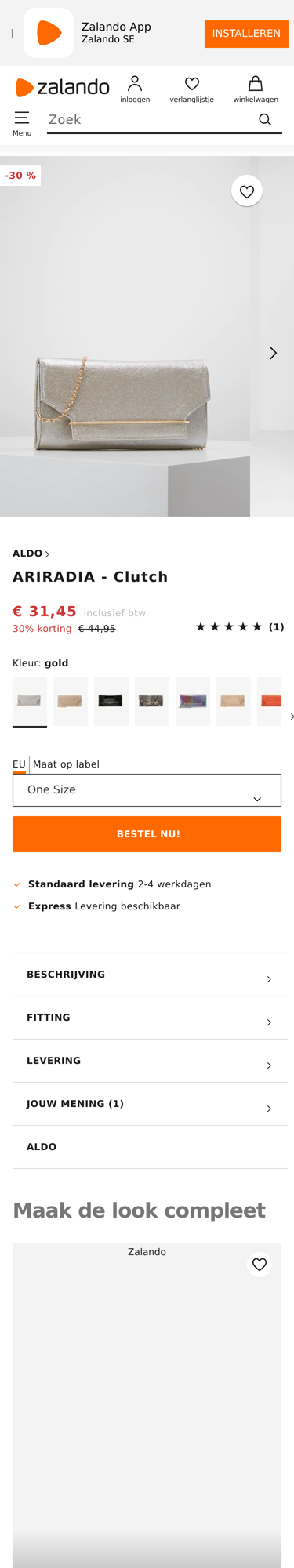}
        \label{fig:subfig3}
    \end{subfigure}
    \hfill
    \begin{subfigure}{}
    \includegraphics[width=0.18\textwidth, trim= 0 38cm 0 0, clip]{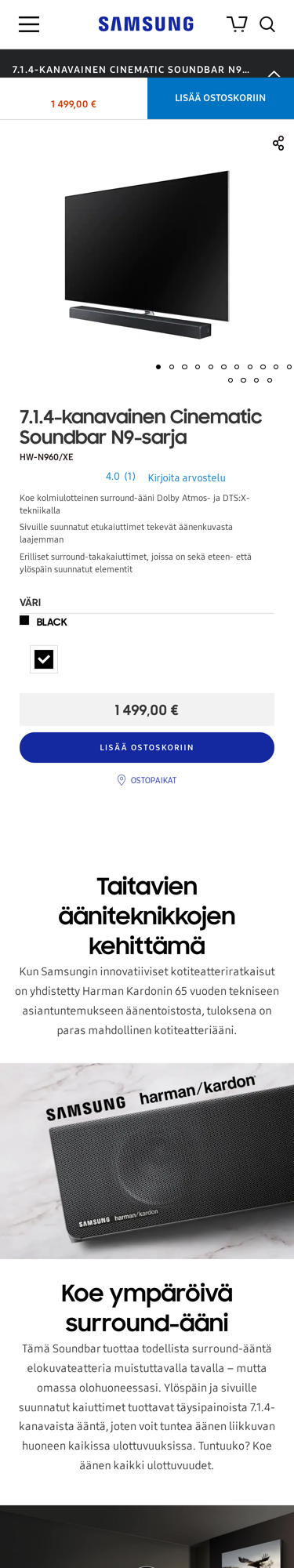}
        \label{fig:subfig4}
    \end{subfigure}
    \caption{Four pages from the Finnish, Dutch and US markets in our dataset.}
    \label{fig:klapp_screenshots}
\end{figure*}

As a significant contribution to the community, we introduce the \emph{\DatasetName}\protect\footnote{Available at \href{https://github.com/klarna/product-page-dataset}{github.com/klarna/product-page-dataset} under the CC BY-NC-SA license.}. Figure \ref{fig:klapp_screenshots} presents screenshots of four rendered pages from our dataset. The \DatasetName\ is designed to emulate a scenario encountered by, for instance, an autonomous shopping assistant. This assistant is tasked with navigating through product pages across a multitude of different websites to extract product information. The goal could be to identify a specific item requested by a user at a given price point, then add this item to the cart and proceed to the checkout page. However, note that in the empirical evaluation carried out in this work, we solely focus on identifying the elements individually. 

Furthermore, the \DatasetName\ dataset comprises over $50,000$ real e-commerce pages from $8,175$ websites, with five manually labeled action and information elements per page (See Figure \ref{dataset:example_page}). To the best of our knowledge, this positions our dataset as the largest available labeled webpage dataset in terms of the number of websites. Additionally, while the majority of public datasets primarily consist of English pages, a unique advantage of \DatasetName\ is its inclusion of pages from eight distinct geographic markets. This allows for the incorporation of text elements written in multiple languages (See Table \ref{tab:dataset:overview}). Finally, recent studies have also leveraged webpage screenshots to explore computer vision research for the web \cite{zheng2024gpt, kumar-etal-2022-cova}. To support future similar efforts, we have developed a complementary dataset that includes screenshots of the rendered pages from our dataset.

 As a starting point for research on the \emph{\DatasetName}, we benchmark the performance of various neural networks on \emph{Web element nomination}, which is a fundamental concept for automating tasks on the web. We define Web element nomination as the process of identifying a unique element from a specific class on a webpage. One of the challenges that makes this task particularly demanding is that among potentially thousands of nodes on a page, only one element belongs to each class. For this purpose, we train several neural networks from three popular families of graph neural networks (GNNs): recurrent GNNs (Tree Long Short-Term Memory networks), convolutional GNNs (GCNs), and attention-based GNNs. Our aim is to compare their performance in web element nomination and identify the most promising GNN family for future research on the \emph{\DatasetName}. We also compare their performance against state-of-the-art (SOTA) methods FreeDOM \citep{Lin_2020} and DOM-Q-NET \citep{jia2018domqnet}, both of which are fairly intricate algorithms. For instance, FreeDOM requires a two-stage training process, and both algorithms utilize information from the entire DOM tree to represent a single element. We discovered that a surprisingly simple GCN, which solely uses local neighbourhood information around each node in the DOM tree, outperforms these SOTA methods. This finding highlights the potential of employing GCNs as building blocks in more advanced web element nomination architectures in future research.

 We subsequently investigate three methodologies aimed at further enhancing the web element nomination performance. Firstly, we explore whether the text present on the page is informative for nominating web elements. Specifically, we use a pre-trained \emph{language model} to create embeddings of the text on the page, which we then add to the set of features. Our findings indicate that models incorporating text in general achieve higher average accuracy. Surprisingly, for certain tasks, the included text proves to be non-informative and, in fact, reduces performance. Secondly, applying LLMs to the entire HTML of realistic pages is unfeasible, partly due to limited context windows but mainly due to the high costs associated with processing a large number of tokens. To address this, we explore an approach where we significantly enhance the performance of the highest-performing GNN model with a post-training refinement step where an LLM is applied to a small subset of the elements of each page. Finally, we address the absence of a standardized training methodology for element nomination \footnote{For comparison, see vertex nomination, as in \cite{fishkind2015vertex, coppersmith2014vertex}. This field typically involves very small numbers of graphs (in this context, webpages), rendering these methods unsuitable here.}. To this end, we propose and evaluate a novel training approach that considerably improves nomination accuracy. 

To summarize, our aim is to lower the entry barriers in this exciting field through the following contributions:

\begin{itemize}
    \item[1.] We introduce the \emph{\DatasetName}\protect\footnote{Available at https://github.com/klarna/product-page-dataset under the CC BY-NC-SA license.}, which consists of $51,701$ manually labeled product pages from $8,175$ e-commerce websites. Suitable for tasks such as element nomination, evaluation of LLMs, or fine-tuning of foundation models. We have also created a complementary dataset with the corresponding screenshots of the webpages for future CV applications.

    \item[2.] We adapt and apply six algorithms from three of the most popular families of GNNs to a web element nomination task, namely recurrent GNNs (Tree Long Short-Term Memory networks), convolutional GNNs (GCNs) and attention based GNNs. We also investigate the performance of SOTA baselines for element nomination. In particular, we identify that \emph{GCN-Mean}, a simple 2-layer GCN, outperforms the \emph{second-best} model by a impressive margin of $9.2$ percentage points measured in average nomination accuracy.
    
    \item[3.]  We explore a post-training nomination refinement step using an LLM. Specifically, we use a trained 2-layer GCN-Mean algorithm to filter out all elements except the top ten for each class on every webpage in the test set. We then use GPT-4 to perform the final nomination based on the local HTML content of these filtered elements. We show that this refinement step substantially enhances nomination accuracy across all explored nomination tasks, resulting in an average increase of $16.82$ percentage points.

    \item[4.] Typically, in element nomination, one would train a model to classify elements and then evaluate the model based on its nomination performance. We refer to this approach as the \emph{Basic Nomination Training Procedure}. We introduce the \emph{Challenge Nomination Training Procedure} (\TrainingName), a novel training method that steers the classification objective towards the nomination objective. Our results demonstrate that \TrainingName\ substantially improves nomination accuracy by approximately 5 percentage points for our best model.
      
    \item[5.] We assess the impact that the text on the pages has on nomination accuracy by utilizing a \emph{language model} to embed the text and integrate it as a feature. Our findings demonstrate that this considerably improves performance in specific nomination tasks, such as when identifying the Buy Button. 
 
\end{itemize}


\begin{table}[!htbp]
\centering
\caption{An overview of the \DatasetName.}\label{tab:dataset:overview}
   \begin{small}
\begin{tabular}{l l c c c c}
        \toprule
        market& language & \# sites & \# pages & median \# nodes\\
        \midrule
DE& German& $2,941$& $16,765$ & $1,391$\\ 
US& English& \textbf{$1,794$}& \textbf{$11,003$} & $1,394$\\ 

GB& English& \textbf{$1,360$}& \textbf{$11,144$} & $1,291$\\ 
FI& Finnish& $1,125$& $5,623$ & $779$\\ 
AT& German& $899$ & $1,316$ & $1,499$\\ 
SE& Swedish& $619$& $4,866$& $1,450$  \\ 
NO& Norwegian& $180$& $852$ & $1,477$\\
NL& Dutch& $130$& $132$ & $1,628$ \\

\midrule
Total  && $8,175$& $51,701$ & $1,308$ \\ 
\bottomrule
\end{tabular}
\end{small}
\end{table}

\section{The \DatasetName}\label{sec:dataset}
\begin{figure}[t]
\centering
\includegraphics[width=0.2\textwidth, trim= 0 7cm 0 0, clip]{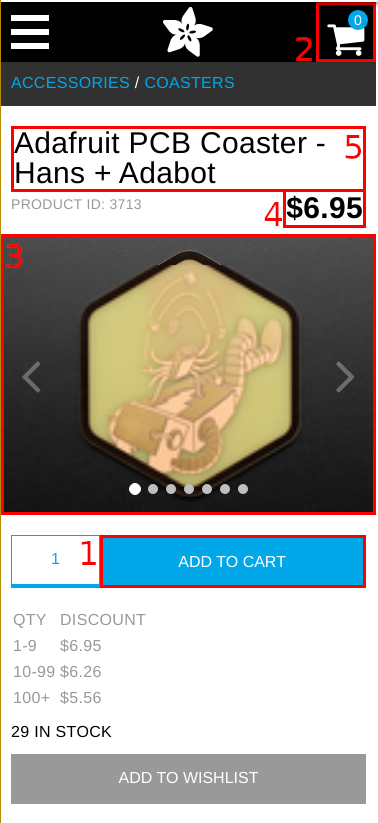}
\caption{A page in our dataset. The labeled elements are surrounded by red boxes:
(1) \emph{buy button},
(2) \emph{cart button},
(3) \emph{image},
(4) \emph{price},
(5) \emph{name}.
}\label{dataset:example_page}
\end{figure}

We collected the \DatasetName\ dataset over several months between $2018$ and $2019$. The dataset comprises $51,701$ product pages from $8,175$ distinct e-commerce merchants. It is conveniently divided into $80\%$ training and $20\%$ test sets, which ensures that pages from the same website do not appear in both the training and test sets. Table~\ref{tab:dataset:overview} presents an overview of the dataset statistics. Each page in our dataset is saved as an MHTML file, which includes all images and assets required to render the page. Furthermore, every page in the dataset has $5$ labeled elements: $2$ corresponding to action elements (\emph{buy button} and \emph{cart button}) and $3$ to information elements (\emph{product price}, \emph{product name}, and \emph{product image}). These labels have been manually annotated by human analysts. In our experiments, we also consider a $6$th, more abstract label: the \emph{subject node}, defined as the lowest common ancestor of all other labels. An example page screenshot from our dataset with marked labels is shown in Figure \ref{dataset:example_page}. Additionally, for convenience, we have created a complementary dataset of page screenshots, suitable for computer vision research.

Moreover, the analysts were given instructions on how to conduct the labeling process, which potentially biased the data collection. First, they were given a list of e-commerce domains that was biased towards popular e-commerce merchants. Then, for every website, they were asked to identify a small number of pages related to physical products. See Table~\ref{tab:dataset:overview} for the specific quantities. They were also instructed to select products that were in stock and to avoid product pages with configuration fields (e.g., color and size). Once on a product page, they labeled elements that corresponded to the product, i.e. \emph{price}, \emph{name}, and \emph{image}, as well as the \emph{cart} and \emph{buy} buttons by clicking on them on the rendered HTML using an in-house developed tool. The tool rendered the pages in a mobile viewport using an iPhone X device agent. Thus mobile versions of pages are over-represented. However, not all domains are rendered differently in mobile browsers.

\section{Element nomination}\label{sec:problem_statement}

\begin{figure*}[t]
  \centering
\begin{tikzpicture}[
    fillednode/.style={ 
    every node,
    font=\footnotesize,
    draw,
    circle,
    align=center,
    minimum size=0.6cm, 
    inner sep=1.75pt,
    fill=C9!69,
    line width=3pt,
    draw=C9,
    very thick,
    text = black!72,
  },
        fillednode2/.style={ 
    every node,
    font=\tiny,
    draw,
    circle,
    align=center,
    minimum size=0.4cm, 
    inner sep=1pt,
    fill=C9!69,
    line width=3pt,
    draw=C9,
    very thick,
        text = black!72,
  },
          fillednode3/.style={ 
    every node,
    font=\tiny,
    draw,
    circle,
    align=center,
    minimum size=0.1cm, 
    inner sep=0.05pt,
    fill=C9!69,
    line width=3pt,
    draw=C9,
    very thick,
        text = black!72,
  }
]

\input{tikz-figures/big_dom_tree}
\input{tikz-figures/classification}
\input{tikz-figures/nomination}

\end{tikzpicture}
 
  \caption{\textbf{Difference between nomination and classification:} To the left, a DOM-tree representation of a webpage is depicted. During the classification process, the Graph Neural Network embedder takes a node and its context, here being its directly neighboring nodes, as input. The output from the embedder is then fed into a classification layer that produces classification scores. In element nomination, the model is applied to every node in the tree. For each label type, the resulting classification scores are ranked across all nodes, and the node receiving the highest rank is nominated for that specific label type. For example, here we see how the node receiving the highest ranking for the \emph{Buy Button} label is nominated as the \emph{Buy Button} element of the page.}
  \label{fig:diff_nom_class}
\end{figure*}

Element nomination plays a crucial role in web automation features, where a common subtask involves accurately selecting a single element from a webpage for a specific action.  In such scenarios, the system often has only one opportunity to perform the task correctly. Incorrect actions, such as clicking the wrong button or filling in the wrong form, could significantly degrade the user experience. Therefore, the ability to precisely identify the correct element before executing any action is vital. 

Unfortunately, there is no established training objective specifically for web element nomination. Instead, the common practice is to train a model for element classification and then assess its nomination performance. The distinction between classification and nomination is illustrated in Figure \ref{fig:diff_nom_class}. During the training phase, a GNN embedder is applied to a small subset of nodes from each page, and the resulting output is subsequently passed to a classifier. After the training phase, the model's performance is assessed based on its ability to nominate elements. Now, the task is to identify a few specific elements on each page. During the evaluation, we nominate elements such as the cart button by applying the trained model to every node on the page to obtain classification scores. These scores are then ranked, and the node with the highest score is nominated as the cart element.

Web element nomination becomes particularly challenging when applied to realistic webpages. A perfect classifier, which always assigns the correct class to every element, would also always nominate the correct element. Hence, the classification objective serves as a proxy for training on the nomination task. However, the pages in our dataset consist of thousands of elements, and among these, there is a single labeled element for each class. For instance, a classifier with $99\%$ accuracy would be expected to incorrectly identify 10 elements as 'add-to-cart' buttons on a page with $1,000$ elements, even though only one true 'add-to-cart' button exists. Consequently, in Section \ref{sec:hardlabels}, we propose an effective, novel training method that boosts the element nomination performance by steering the classification training objective towards the nomination objective.

\section{Related work}\label{sec:related-work}
In the literature, there is an unfortunate scarcity of datasets suitable for web element nomination. In this section, we walk through and compare existing datasets to the \DatasetName. 




The minimalistic World-of-bits environment \citep{shi2017world}, known as Mini-WoB, is useful for task-solving on the Web. However, the pages in Mini-WoB are simple compared to sites found on the Internet. A page in Mini-WoB might, for instance, consist of just a single form and a submit button. In contrast, the \DatasetName\ comprises manually annotated clones of real webpages, with the median page featuring $1,308$ elements and thousands of lines of code. The recently proposed WebShop \citep{yao2022webshop} is another simulated interactive environment where agents can learn to navigate an e-commerce store and make purchases based on text instructions. WebShop includes 1.18 million products and $12,087$ text instructions. While it effectively emulates an e-commerce store, its primary limitation is that it comprises pages from a single website.



Additionally, there are datasets related to ours that are used for web content extraction. Perhaps the most well-known dataset in this domain is the SWDE dataset~\citep{SIGIR-2011-HaoCPZ}, which contains 124,291 real webpages from 80 different websites across 8 verticals, such as online book and camera stores. Each webpage in the SWDE dataset is labeled with 3-5 attributes, depending on the vertical. For instance, in the book vertical, the focus is on identifying the book's title and author, while in the automotive vertical, it involves extracting the model and price of the vehicle. Previous studies \citep{Lin_2020,DBLP:journals/corr/abs-2101-02415} have noted that test accuracy increases with the number of sites included in the training set for each vertical. However, the SWDE dataset is limited by having only 10 sites per vertical, thereby capping the potential gains. In comparison, our dataset consists of $8,175$ diverse e-commerce websites, and thus provides a much more comprehensive representation of a vertical.
The \DatasetName\ also includes labels that introduce a broader range of learning challenges. Whereas the SWDE dataset contains solely labeled leaf nodes, which are typically identifiable based on local information, our dataset includes labels where local information alone is insufficient for accurate identification. For instance, elements like the \emph{price} and \emph{name} are primarily defined by local information, making their identification similar to standard web data extraction tasks. Conversely, nominating the \emph{buy} button likely requires both local and contextual information. Lastly, the \emph{subject node} has no meaning on its own and can only be represented based on its context. Finally, the SWDE dataset consists of only English pages, while our dataset includes pages with text written in multiple languages.

Nevertheless, a recent large-scale webpage dataset that signifies a substantial advancement over previous works is Mind2Web \cite{deng2023mind2web}. Mind2Web encompasses 137 websites across 31 diverse domains, such as Music, Airlines, Housing, and Social Media. Consequently, it effectively provides the opportunity to measure generalizability across various tasks and domains. We believe that our dataset complements Mind2Web in this aspect, as the \DatasetName\ offers the possibility of measuring generalizability for the same task across a vast number of websites.

Overall, we firmly believe that the diverse range of webpage structures and labels in our dataset represents a significant advancement over current SOTA datasets for web element nomination. However, we also emphasize the invaluable role of having access to a variety of datasets, as this enables testing the generalization capabilities of algorithms across different contexts.

\section{Embedders for DOM Elements}\label{sec:applications}

We now describe the GNN embedders that we train together with a classification layer (see Fig \ref{fig:diff_nom_class}) on our dataset. The embedders represent three GNN families, specifically recurrent, convolutional and attention-based networks. We now detail and then evaluate these embedders in the simulation section. 

\textbf{GCN-Mean}: we first consider a multi-layer GCN model inspired by GraphSage~\citep{10.5555/3294771.3294869} and PinSage~\citep{ying2018graph}. Each convolutional layer $\ell$, computes the representation of the current node, $v$, as $\bz_v^{(\ell)}$ based on the local encoding and an average encoding 
$\bh^{(\ell)}$ of its immediate neigborhood $\mathcal{N}(v)$. 
The representation is obtained according to the following:
\begin{equation}
    \begin{split}
    \bh^{(\ell)}_{v} &= \textrm{AVG}\del[1]{\cbr[0]{\textrm{ReLU}(\bV_l \bz_u^{(\ell-1)} + \bb_l),\; u\in \mathcal{N}(v)}}, \\
    \bk^{(\ell)}_v &= \textrm{ReLU}\del[1]{\bW_l \cdot \textrm{CONCAT}(\bz_v^{(\ell-1)}, \bh^{(\ell)}_{v}) + \bw_\ell},\\
    \bz^{(\ell)}_v &= \bk^{(\ell)}_v / \enVert[1]{\bk^{(\ell)}_v}_2,
    \end{split}
\end{equation}

where $\bV_l$, $\bW_l$, $\bb_l$ and $\bw_l$ are trainable weight matrices for convolution layer $\ell$ and $\bz_i^{(0)} =
\bx_i^{(0)}$ for all $i$. We use the same dimension for the encodings $\bh_v^{(\ell)}$ and $\bk_v^{(\ell)}$. Encoding size, representation size, and total number of layers are tunable hyperparameters. By stacking $K$ convolution layers, information can be propagated from $K$-hop neighborhoods.

\textbf{TransformerEncoder}: This model consists of a single-layer multi-headed attention encoder stack~\citep{vaswani2017attention} fed with a
sequence consisting of the features $\bx_v$ of the local node, $v$, stacked with the features $\bx_u$ of its neighbors $u\in\mathcal{N}(v)$. The embedding $\bz_v$ of node $v$ is then extracted as the first element (here, at index $0$) of the sequence $\bH_v$ computed by the transformer encoder stack:

\begin{equation*}
\begin{split}
        \bH_v &= \textrm{TransformerEncoder}\!\del[1]{\textrm{STACK}\del[1]{\bx_v,\, \{\bx_u,\; u\in\mathcal{N}(v)\}}},\\
        \bz_v &= \bH_v[0].\notag
\end{split}
\end{equation*}
In our implementation, the number of heads in the encoder and the embedding size are tunable hyperparameters. As the
elements in the DOM tree do not have an evident natural linear ordering, we did not use positional encodings. In fact, how to encode positional information in trees is the focus of ongoing research.

\textbf{Tree-LSTMs}: We consider four tree-based LSTM models. They all revolve around a standard LSTM cell~\citep{hochreiter1997long} and are characterized by how the input sequence is constructed from the DOM
tree.

In the top-down model (LSTM-TD), the embedding $\bz^D_v$ of node $v$ is the last element of the LSTM encoding of the path from the
root to node $v$: each unit receives hidden and cell states from its parent, according to
\begin{equation*}
    \bh^D_v, \bc^D_v = \textrm{LSTM}(\bx_v, \bh^D_{\textrm{Parent}(v)}, \bc^D_{\textrm{Parent}(v)}),\qquad
    \bz^D_v = \bh^D_v.
\end{equation*}
where $\textrm{LSTM}(x,h,c)$ denotes a standard LSTM cell with input $x$, hidden state $h$ and cell state $c$. Cell and
hidden states were initialized as zero for the root node.

In the bottom-up model (LSTM-BU), we consider the child-sum tree-LSTM algorithm~\citep{tai2015improved}. Here, the sequence is
defined recursively from the leaves to each node $v$. At each node, the hidden state is the sum of the hidden states of
the children, and the cell state is computed based on the cell states of the children, with one forget gate per
child state.
\begin{equation*}
\begin{split}
\bh^U_v, \bc^U_v &= \textrm{ChildSumLSTM}\!\del[1]{\bx_v, \cbr[0]{\bh^U_u,\, u \in \mathcal{C}(u)},\, \cbr[1]{\bc^U_u,\, u\in \mathcal{C}(u)}}, \\
\bz^U_v &= \bh^U_v.
\end{split}
\end{equation*}

 To account for more context when computing the element embeddings, we combined the bottom-up and top-down
 approaches into a bidirectional model (LSTM-Bi) where the embeddings are the concatenation of the two previous models' embeddings, $\bz^B_v = \textrm{CONCAT}(\bz^D_v,\, \bz^U_v)$.  In this case, the weights in the top-down and the bottom-up components are trained simultaneously.

 Finally, we considered a global bidirectional model (LSTM-BiE) that computes node representations based on the whole DOM-tree~\citep{cook2019learning}. This method has three steps. First, a bottom-up representation $\bz^U_{v}$ is calculated
for each node $v$. Second, this representation is then used as the input to a top-Down architecture that gives a
representation $\bz^E_v$ for each node $v$:
\begin{equation*}
    \bh^E_v, \bc^E_v = \textrm{LSTM}(\bz^U_v, \bh^E_{\textrm{Parent}(v)}, \bc^E_{\textrm{Parent}(v)}),\qquad
    \bz^E_v = \bh^E_v.
\end{equation*}
Third, the two representations are concatenated into a final representation for the node $\bz^{BiE}_v = \textrm{CONCAT}(\bz^E_v, \bz^U_v)$. Also, in this case, the weights in both components are trained simultaneously.

\textbf{GCN-GRU}: This model is inspired by the local embedding module in the DOM-Q-Net algorithm~\citep{jia2018domqnet}. Here, the
encoding of node $v$ is computed by feeding a Gated Recurrent Unit (GRU) with the local features of the node and an
average encoding of the neighborhood of the node; then, the embedding is computed with a transformation of the local
encoding $\bh_v$ and the input features:
\begin{equation*}
    \begin{split}
        \bh_v &= \textrm{GRU}\!\del[1]{\bx_v, \, \mathrm{AVG}\!\del[1]{\cbr[0]{\bV \bx_u + \bb,\,u\in \mathcal{N}(u)}}}, \\
        \bz_v &= \bW\cdot \textrm{CONCAT}(\bx_v,\,\bh_v) + \bw,
    \end{split}
\end{equation*}
where $\bV$, $\bb$, $\bW$, and $\bw$, are trained together with the parameters of the GRU.

\textbf{FreeDOM}: We adapt the FreeDOM architecture \citep{Lin_2020}. This is a two-stage method where a neural
network first computes node representations based on local and contextual information (using both text and
HTML), and then uses representation distances and semantic relatedness between node pairs in the second stage. In the original implementation, \citep{Lin_2020} perform a filtering step where templates common to multiple pages are identified and used to reduce the number of nodes that the model considers. On our dataset, this is not possible since the pages generally do not come from the same set of websites. That being said, FreeDOM may be at a disadvantage compared to the other models because it does not perfectly suit our application or dataset. However, the authors set out to learn site-invariant feature representations and found that FreeDOM could generalize well to unseen websites. 


\section{Element Nomination Evaluation}\label{sec:experiments}
As a starting point for research on the \DatasetName\ we now compare various GNNs on the element nomination task. We first present the common setup of the experiments. Then, in Section \ref{exp:eval_pred_acc}, we present the results of the Basic Training Procedure. Later, in Section \ref{sec:llmexperiment}, we employ the best-performing model from the initial evaluation to filter the webpages in our datasets. We then apply GPT-4 to the HTML of the remaining elements to further improve nomination. Finally, in Section \ref{sec:hardlabels}, we illustrate how the CNTP training scheme enhances nomination performance.

\textbf{\quad Training Procedures.} In our experiments, we train the GNN embedders together with a classification layer (see Fig \ref{fig:diff_nom_class}). Thus, as described in Section \ref{sec:problem_statement}, each model learns to classify elements during training. More specifically, an embedder takes a node with some neighborhood as input, and this output is then followed by a single-layer fully connected neural network, which acts as a classifier. In the \emph{Basic Nomination Training Procedure}, presented in Section \ref{exp:eval_pred_acc}, every model classifies $5$ labeled and $10$ randomly sampled unlabeled DOM elements from each page in the training set. In the \emph{Challenge Nomination Training Procedure}, presented in Section \ref{sec:hardlabels}, we conduct an additional experiment where we significantly improve nomination accuracy by selecting negative examples that the models confidently misclassify. Furthermore, the classification gives rise to a cross-entropy training objective that is minimised with the Adam optimiser.\newline
\textbf{\qquad  Embedders.} We consider embedders from three important GNN families, specifically recurrent, convolutional and attention-based networks. We consider LSTM-based embedders (LSTM-TD, LSTM-BU, LSTM-Bi, LSTM-BiE), the mean-pooled GCN model (GCN-Mean), the GRU based GCN inspired by DOM-Q-Net (GCN-GRU), and the transformer-encoder (TE). For comparison, we also present the results of the SOTA FreeDOM architecture, which we train on two different sets of features. First, we use the original features defined in~\citep{Lin_2020}, then we specify an extended version (FreeDOM-ext) which, in addition, uses the style features used by our other models (see further down). We also implement a 2-layer, \textit{Fully Connected Network (FCN)} that uses the same input features as the other models and thus acts as a context-oblivious baseline. With the trained model we then evaluate how accurately the GNNs can nominate the six labels described in Section \ref{sec:dataset}.\newline
\textbf{\hspace{2cm} Features.} For each node's \emph{local features}, we define a set of \emph{style features} which consists of the bounding box of the elements as rendered on the page, font weight, font size, number of images contained within the subtree, visibility, and HTML tag. For the FreeDOM-specific, pre-trained, NLP features described in~\citep{Lin_2020}, we use the Spacy\footnote{v2.3.5 \textit{en\_core\_web\_sm}  \url{https://spacy.io/models/en} (MIT License)}\citep{spacy} library. To gauge the impact of textual features on the other models, we perform an experiment where we augment the features with pre-trained embeddings computed using the~\emph{Universal Sentence Encoder}\footnote{Version 4, \url{https://tfhub.dev/google/universal-sentence-encoder/4} (Apache-2.0 License)}~\citep{cer-etal-2018-universal}. FreeDOM's implementation stays the same in this experiment as it already uses text.\newline
\textbf{\hspace{2cm} Hyper-Parameters.} For the experiments, we ensured that all hyper-parameters were optimized for this setting. For FreeDOM, we used the hyper-parameters provided by the original paper since our setting is very similar. For the TreeLSTM-based algorithms, we used the results of the hyper-parameter tuning in \citep{cook2019learning}. We then performed hyper-parameter tuning on GCN-Mean, GCN-GRU, FCN and the TransformerEncoder. We tuned our method and these other particular algorithms since we have adapted and applied them to a substantially different setting than what they were designed for. In the tuning procedure, we performed a grid-search over the values in \textsc{hyperparameter\_grids} folder\protect\footnote{The code, along with all the hyperparameters, is available at https://github.com/klarna/product-page-dataset.} using the \textsc{Ray Tune} \citep{liaw2018tune} library. 
\textbf{\hspace{2cm} Evaluation setup.} After training the models, we evaluate them by measuring their classification and nomination performance in two scenarios: \textbf{1)} only using style features, and \textbf{2)} using text embeddings in addition to the style features. 







\begin{table*}[hb]
\centering
\caption{The nomination accuracy of each model on each label without (left) and with text (right) added to the element features.}\label{exp:nomstats}
\begin{adjustbox}{width=\textwidth}

\begin{scriptsize}
\begin{tabular}{l|c|c|c|c|c|c|c|c|c|c|c|c|c|c|c}
\toprule
& \multicolumn{2}{c|}{buy btn} & \multicolumn{2}{c|}{cart btn} & \multicolumn{2}{c|}{main img} & \multicolumn{2}{c|}{name} & \multicolumn{2}{c|}{price} & \multicolumn{2}{c|}{subj} & \multicolumn{2}{c|}{avg} \\
\midrule
Model & w/o & w/ & w/o & w/ & w/o & w/ & w/o & w/ & w/o & w/ & w/o & w/ & w/o & w/ \\
\midrule
LSTM-Bi* & .72 & - & .57 & - & .47 & - & .69 & - & .48 & - & .66 & - & .60 & - \\
LSTM-BiE*& .73 & - & .60 & - & .50 & - & .72 & - & .55 & - & .70 & - & .63 & - \\
GCN-GRU & .77 & .83 & .57 & .62 & .41 & .38 & .69 & .79 & .50 & .60 & .45 & .53 &  .57 & .63 \\
FreeDOM** & - & .29 & - & .06 & - & .04 & - & .57 & - & .23 & - & .35 & - & .26 \\
FreeDOM-ext** & - & .84 & - & .59 & - & .31 & - & .84 & - & \textbf{\bfseries .72} & - & .46 & - & .63 \\
LSTM-TD & .71 & .85 & .50 & .51 & .42 & .46 & .70 & .76 & .52 & .58 & .59 & .34 & .57 & .58 \\
LSTM-BU & .73 & .89 & .57 & .65 & .45 & .44 & .69 & .74 & .46 & .53 & .66 & .56 & .59 & .64 \\
TE & .69 & .81 & .43 & .46 & .25 & .34 & .73 & .79 & .60 & .62 & .55 & .62 & .54 & .61 \\
FCN & \textbf{\bfseries{.79}} & .86 & .53 & .62 & .32 & .27 & .70 & \textbf{\bfseries{.85}} & .58 & .61 & .43 & .39 & .56 & .60 \\
GCN-Mean & .78 & \textbf{\bfseries .91} & \textbf{\bfseries{.62}} & \textbf{\bfseries .67} & \textbf{\bfseries .61} & \textbf{\bfseries{.48}} & \textbf{\bfseries .78} & .81 & \textbf{\bfseries .66} & .66 & \textbf{\bfseries .84} & \textbf{\bfseries .86} & \textbf{\bfseries .71} & \textbf{\bfseries .73} \\
\bottomrule
\end{tabular}
\end{scriptsize}

\end{adjustbox}
\end{table*}
\subsection{Basic Training Procedure}\label{exp:eval_pred_acc}
\textbf{Setup.} In this model comparison we trained them on a subset of the \DatasetName\ containing $10,000$ pages from English markets. We did not train the models on the entire dataset because the LSTM models are very computationally expensive. We only used pages from English markets to facilitate the use of a pre-trained text embedder based on a well-known approach. Also, it offers a more level playing field for the SOTA FreeDOM architecture, since it relies on page text and was originally applied to English websites.\newline
\textbf{Evaluation.} After training, the models are evaluated based on nomination performance. As described in Section \ref{sec:problem_statement}, with the exception of FreeDOM, each trained model classifies all nodes on every webpage in the test set. The nodes are then ranked for each of the classes \emph{price}, \emph{name}, \emph{image}, \emph{buy button}, \emph{cart button} and \emph{subject node} and the node with the highest probability for each class is nominated for that class. The answer is correct if the nominated element is the unique element with the ground truth label. For FreeDOM, we instead followed the protocol as described in \citep{Lin_2020}. The evaluation is performed on $6,706$ pages from merchants not present in the training set.

\textbf{Results.} The nomination accuracy results can be found in Table~\ref{exp:nomstats}. Note that here precision and recall both equal accuracy as the algorithms make exactly one guess on each page, and there is one correct element per class. 



From the results with only style features (no text features) in Table \ref{exp:nomstats}, the first thing we notice is that GCN-Mean achieves the highest average accuracy by a wide margin ($6.8$ p.p.). Despite its relative simplicity, GCN-Mean consistently scores among the best for all tasks explored here. This suggests that a large proportion of the context that is relevant for element nomination is concentrated in an element's vicinity. Among the runner-ups, LSTM-BiE stands out, as it shows similar performance to FreeDOM-ext (using text), while GCN-GRU, LSTM-TD, LSTM-Bi and LSTM-BU all show similar average performance. It is worth noting that the LSTMs require substantially more compute than GCN-Mean, GCN-GRU and FreeDOM. Also, FreeDOM requires two separate training stages. 

When we add text features, GCN-Mean still has the highest average accuracy, this time by an even wider margin. Though LSTM-BiE was not trained in with text due to computational constraints, its performance remains competitive with all other algorithms except GCN-Mean when trained on the style features. This further solidifies our intuition that straightforward graphical algorithms such as GCN-Mean and LSTM-BiE should be explored further as building blocks in Web element nomination algorithms.


One important benefit of LSTM-BiE and GCN-Mean over FreeDOM, SimpDOM and the WebFormer is that they remain competitive without text features. In practice, this allows us to deploy a single model that can be used across webpages, regardless of the language used on the page.

\subsubsection{The Impact of Text and Contextual Information on Accuracy}\label{sec:context_important}

The experiments show that more context helps performance on certain tasks. For all tasks, we use the performance of the FCN as an indicator of the importance of context. More specifically, the gap between the best model and the FCN gives us a lower bound on the gain that we can achieve by including contextual information. With this in mind, it is possible to identify tasks where contextual information is more important. For example, the \emph{subject node} on its own does not contain much information. Instead, the \emph{subject node} is defined by its relationship to other nodes in the tree, and, therefore, we need to look at its context to identify it. For other tasks, context is less important. For instance, the \emph{buy button} seems to contain sufficient information to be correctly identified based on its local features. Another interesting observation is that LSTM-BiE outperforms LSTM-Bi on all tasks; this indicates that processing all context from the page improves accuracy.

One important finding is that algorithms that try to exploit text perform worse when this information is irrelevant to the task. We can observe this degradation when GCN-Mean, GCN-GRU, and even FCN attempt to use text to nominate the image node. LSTM-TD performance drops when text is added to its feature set since it looks at the text in nodes above the \textit{subject node}. The \textit{subject nodes} are usually close to the root of the DOM tree; hence, it is unlikely that nodes in the considered sequence even have text. This variety in performance suggests that we need varying degrees of contextual information to nominate different label types. This also suggests that the \DatasetName\ presents a comprehensive set of challenges that not all algorithms can easily overcome.

\subsection{LLM Refinement Step}\label{sec:llmexperiment} \textbf{Setup.} The refinement step was initiated using the Two-Layer GCN-Mean algorithm, trained with the Basic Training Procedure, to filter out all non-informative elements from the test set. This filtering was performed by ranking the elements by their classification scores and retaining the ten highest-ranked elements. The raw HTML of these elements, combined with their bounding boxes, was then used to formulate a query where the task was to select the single labeled element from the set. We then presented this query to GPT-4, which provided an answer. This experiment was constrained to $100$ randomly sampled English pages from the test set, due to budget limitations.

\textbf{Results.} From the results shown in Figure \ref{fig:nomination_refine_llm}, it is evident that the LLM refinement step plays a crucial role in enhancing the final nominations. This procedure significantly boosts the nomination performance across all tasks, despite solely relying on local element content. The improvement is particularly noticeable for the product image element, indicating that the HTML contains substantial information for this specific task. However, there is still room for improvement, which could potentially be achieved by incorporating screenshots of the rendered pages from our dataset.

\begin{figure}[tb]
\tikzstyle{f0}=[ ybar, fill=C6!60, draw = C6]
\tikzstyle{f4}=[ybar, fill=C0!60, draw = C0]
\setlength{\mybarwidth}{9pt}
\centering
\begin{tikzpicture}[scale=1]
    \begin{axis}[
        ybar,
        ymajorgrids = true,
        height = 0.7\linewidth,
        width =  \linewidth,
        bar width=0.8\mybarwidth,
        symbolic x coords={Buy button, Cart, Image, Name, Price, Avg},
        xtick=data,
        x tick label style={rotate=45, yshift=-1pt, font=\footnotesize}, 
        ylabel = {Nomination accuracy},
        enlarge x limits=0.15,
        y tick label style={/pgf/number format/skip 0.=true},
        legend style={at={(0.5,-0.3)}, anchor=north,legend columns=-1},
    ]

    \addplot+[ybar, f0] coordinates {(Buy button, 0.8474576271) (Cart, 0.6315789474) (Image, 0.4603174603) (Name, 0.8169014085) (Price, 0.6153846154) (Avg, 0.6763754045)};
    \addplot+[f4, ybar] coordinates {(Buy button, 0.9672131148) (Cart, 0.8421052632) (Image, 0.6507936508) (Name, 0.9428571429) (Price, 0.8125) (Avg, 0.8445945946)};

    \legend{GCN-Mean, GCN-Mean \& GPT-4}
    \end{axis}
\end{tikzpicture}

\caption{Substantially Enhanced Nomination Accuracy by Combining GCN-Mean with GPT-4: The performance of the GCN-Mean algorithm alone versus its performance when augmented with a GPT-4-based refinement step. In the enhanced scenario, the GCN-Mean algorithm initially filters the dataset to retain only the top $10$ elements with the highest classification scores on each page. Subsequently, GPT-4 undertakes the final nomination step, based on the local HTML content of these selected elements. This refinement step significantly boosts nomination accuracy across all tasks, resulting in an average increase of $16.82$ percentage points. }\label{fig:nomination_refine_llm}
\end{figure}

\subsection{Bridging the \emph{classification} and \emph{nomination} objectives}\label{sec:hardlabels}
We propose a novel two-stage training approach called \TrainingName\, which is similar to \cite{ben2021multi} and is inspired by the field of Positive-Unlabelled (PU) Learning (see \cite{bekker2020learning} for a survey). A straightforward approach for nominating elements for a class is to classify and then rank the classification scores of all elements on a page. Using this procedure, we can train on the classification objective, which means that we only need to train on a small subset of the elements from each page in the training set. During training, we then periodically rank the elements on the page and add the confusing elements to the training set. A confusing element is an unlabeled element that receives a very high score for a particular label. This increases the likelihood that the true labelled elements appear at the top of the ranking during the nomination phase.
The parameters of this method are the number of unlabelled elements, $M$, to be included in training from each page at random, the number of epochs $T$ and the number of additional elements $K$, considered \emph{hard} by the model in the nomination task after training for $T$ epochs. The pseudo-code for this training data augmentation step is presented in Algorithm \ref{algo1}.

\begin{algorithm}
	\caption{Hard-Example Augmentation}\label{algo1}
	\begin{algorithmic}[1]
	\State Inputs: $M, K, \epsilon, \mathcal{P}$ (the page dataset)
	\State $H_P := \varnothing, \forall P\in \mathcal{P}$
	\State $\mathrm{nn\_model} := \mathrm{model\_init}()$
		\For {$epoch = 1,2,\ldots$}
		    \State $S_{train} := \varnothing$
			\For {\textbf{each} webpage $P \in \mathcal{P}$}
			    \State $P\triangleq (V, E)$, $V\triangleq V_{labelled}\cup V_{unlabelled}$
				\State $S_P := \mathrm{UniformRand}(V_{unlabelled}, \mathrm{n_{samples}}=M)$ 
				\If {$epoch = K$} 
				
				    \For {$l:=1,\dots, L$}
				    \State $preds = \{\mathrm{nn\_model}(v)[l], \forall 
				    \mathrm{elements}\ v\in V\})$
				    \State $H_P := H_P\cup\mathrm{rank}(preds)[:K]$
				    \EndFor
				\EndIf
				\State $S_{train} := S_{train}\cup S_P \cup V_{labelled} \cup H_P$ 
				
			\EndFor
			\State Train nn\_model for $1$ epoch on dataset $S_{train}$.
		\EndFor
	\end{algorithmic} 
\end{algorithm}

\begin{figure*}[t]

\begin{minipage}[b]{0.49\textwidth}
\centering
\tikzstyle{s0}=[very thick, C6]
\tikzstyle{s1}=[very thick, C7]
\tikzstyle{s2}=[very thick, C8]
\tikzstyle{s3}=[very thick, C9]
\tikzstyle{s4}=[very thick, C0]
\tikzstyle{s5}=[very thick, C1]
\tikzstyle{s6}=[very thick, mark=*, C6]
\tikzstyle{s7}=[very thick, mark=*, C7]
\tikzstyle{s8}=[very thick, mark=*, C8]
\tikzstyle{s9}=[very thick, mark=*, C9]
\tikzstyle{s10}=[very thick, mark=*, C0]
\tikzstyle{s11}=[very thick, mark=*, C1]

\pgfplotstableread[col sep=comma]{figure-data/fc_hardlabels_augment_over_time_merged.csv}\data
\begin{tikzpicture}[outer sep = 0pt]
    \begin{axis}[
        ylabel={\#participants},
        width  = \columnwidth,
        height = 0.9*4.5cm,
        ymajorgrids = true,
        ylabel = {Validation Error},
        xlabel = {\# of batches},
        scaled y ticks = false,
        xmin = 29824,
        xmax = 596499,
        ymin=0.33,
        ymax=0.45,
        yticklabel style={/pgf/number format/skip 0.=true},
    ]

    \addplot[s1] table[x index=0,y=49-tag-val] {\data};
    \addplot[s5] table[x index=0,y=54-tag-val] {\data};
    \addplot[s2] table[x index=0,y=50-tag-val] {\data};
    \addplot[s0] table[x index=0,y=51-tag-val] {\data};
    \addplot[s3] table[x index=0,y=52-tag-val] {\data};
    \addplot[s4] table[x index=0,y=53-tag-val] {\data};
    
    \end{axis}%
\end{tikzpicture}%
\end{minipage}%
\hfill
\begin{minipage}[b]{0.49\textwidth}
\centering
\tikzstyle{s0}=[very thick, C6]
\tikzstyle{s1}=[very thick, C7]
\tikzstyle{s2}=[very thick, C8]
\tikzstyle{s3}=[very thick, C9]
\tikzstyle{s4}=[very thick, C0]
\tikzstyle{s5}=[very thick, C1]
\tikzstyle{s6}=[very thick, mark=*, C6]
\tikzstyle{s7}=[very thick, mark=*, C7]
\tikzstyle{s8}=[very thick, mark=*, C8]
\tikzstyle{s9}=[very thick, mark=*, C9]
\tikzstyle{s10}=[very thick, mark=*, C0]
\tikzstyle{s11}=[very thick, mark=*, C1]

\pgfplotstableread[col sep=comma]{figure-data/gcn_hardlabels_augment_over_time_merged.csv}\data
\begin{tikzpicture}[outer sep = 0pt]
    \begin{axis}[
        width  = \columnwidth,
        height = 0.9*4.5cm,
        ymajorgrids = true,
        xlabel = {\# of batches},
        scaled y ticks = false,
        xmin = 29824,
        xmax = 596499,
        ymin=0.18,
        ymax=0.31,
        yticklabel style={/pgf/number format/skip 0.=true},
    ]

    \addplot[s1] table[x index=0,y=value_40-tag-val] {\data};
    \addplot[s5] table[x index=0,y=value_42-tag-val] {\data};
    \addplot[s2] table[x index=0,y=value_44-tag-val] {\data};
    \addplot[s0] table[x index=0,y=value_45-tag-val] {\data};
    \addplot[s3] table[x index=0,y=value_48-tag-val] {\data};
    \addplot[s4] table[x index=0,y=value_47-tag-val] {\data};
    
    \end{axis}%
\end{tikzpicture}%
\end{minipage}

\begin{tikzpicture}
\begin{axis}[
    hide axis,
    legend style={
        font = \small,
        legend columns=3, 
        at={(0.5,-0.3cm)}, 
        anchor=north,  
    },
    xmin=0,
    xmax=1,
    ymin=0,
    ymax=1
]

\addplot[s1] coordinates {(0,0)};
\addlegendentry{Every 50 epochs}
\addplot[s5] coordinates {(0,0)};
\addlegendentry{At epoch 50};
\addplot[s2] coordinates {(0,0)};
\addlegendentry{At epoch 100};
\addplot[s0] coordinates {(0,0)};
\addlegendentry{At epoch 150};
\addplot[s3] coordinates {(0,0)};
\addlegendentry{At epoch 200};
\addplot[s4] coordinates {(0,0)};
 \addlegendentry{Never (T=$\infty$)};
\end{axis}
\end{tikzpicture}

\caption{Effect on average validation error from performing the augmentation step at different times, consistently and not at all for FCN (left) \& GCN-Mean (right).}
\label{fig:aug_over_time}
\end{figure*}

A description of the procedure goes as follows. At the start of every epoch, we start from an empty training set of elements $S_{train}=\varnothing$. Then, for every webpage $P\triangleq (V, E)$, $V$ being the set of elements, in the set of pages $\mathcal{P}$ allocated for training, we uniformly sample a set $S_P = \{v_i\in V_{unlabelled}: i=1,\dots, M\}$ of $M$ unlabelled elements, and add $S_P$ to the classifier's training set, along with all $L$ labelled elements: $S_{train} = \bigcup_{P\in\mathcal{P}} S_P \cup V_{labelled}$. We then proceed to train the model for one epoch. 
At epoch $T$, we evaluate the model on the nomination task on the entire set $\mathcal{P}$ and initialize the sets of \emph{hard} training examples $H_P=\varnothing,\ \forall P\in\mathcal{P}$. Then, $\forall P\in \mathcal{P}$, we add the top $K$ unlabelled elements in the proposed ranking for each of the $L$ labels to $H_P$. It is sufficient for this step to only be done once. For all future epochs, we always consider these elements as additional training examples in the classification task $S_{train} = \bigcup_{P\in\mathcal{P}} H_P \cup S_P\cup V_{labelled}$. This means that we train on $M+L\times K$ unlabelled elements and all the labelled ones. 

This training procedure has two main benefits: 1) it uses only a small fraction of the total elements in the training set, and 2) it provides a bridge between the nomination and the classification tasks. The former benefit allows us to avoid the impractical computational cost of training to rank all examples. Also, in this setting, we only care about the ranking of the top candidates and not about elements that we confidently identified as uninteresting. We achieve the latter benefit by training on more unlabelled elements that the model finds confusing, i.e. elements that are likely to appear high in the nomination ranking. We want to learn to better distinguish between the ground truth elements and the uninteresting elements that receive high scores. 

\begin{figure*}[tb]

\begin{minipage}[b]{0.49\textwidth}
\centering
\tikzstyle{sc0}=[very thick, C6]
\tikzstyle{sc1}=[very thick, C7]
\tikzstyle{sc2}=[very thick, C8]
\tikzstyle{sc3}=[very thick, C9]
\tikzstyle{sc4}=[very thick, C0]
\tikzstyle{sc5}=[very thick, C1]
\tikzstyle{s6}=[very thick, C6]
\tikzstyle{s7}=[very thick, C7]
\tikzstyle{s8}=[very thick, C8]
\tikzstyle{s9}=[very thick, C9]
\tikzstyle{s10}=[very thick, C0]
\tikzstyle{s11}=[very thick, C1]

\begin{tikzpicture}
\begin{axis}[
        width  = \columnwidth,
        height = 0.9*4.5cm,
        ymax=0.65,
        scaled y ticks = false,
        ylabel = {Avg. Confidence Gap},
        xlabel = {\# of batches},
        ymajorgrids = true,]
\addplot[sc0] table [x=Step, y=Value, col sep=comma] {figure-data/hard-labels/confidence_gap_no_aug/run-version_15_confidence_gap_addtocart-tag-confidence_gap.csv};
\addplot[sc1] table [x=Step, y=Value, col sep=comma] {figure-data/hard-labels/confidence_gap_no_aug/run-version_15_confidence_gap_cart-tag-confidence_gap.csv};
\addplot[sc2] table [x=Step, y=Value, col sep=comma] {figure-data/hard-labels/confidence_gap_no_aug/run-version_15_confidence_gap_mainpicture-tag-confidence_gap.csv};
\addplot[sc3] table [x=Step, y=Value, col sep=comma] {figure-data/hard-labels/confidence_gap_no_aug/run-version_15_confidence_gap_name-tag-confidence_gap.csv};
\addplot[sc4] table [x=Step, y=Value, col sep=comma] {figure-data/hard-labels/confidence_gap_no_aug/run-version_15_confidence_gap_price-tag-confidence_gap.csv};
\addplot[sc5] table [x=Step, y=Value, col sep=comma] {figure-data/hard-labels/confidence_gap_no_aug/run-version_15_confidence_gap_subjectnode_prod-tag-confidence_gap.csv};
\end{axis}
\end{tikzpicture}
\hfill
\end{minipage}%
\begin{minipage}[b]{0.49\textwidth}
\centering
\begin{tikzpicture}
\begin{axis}[
        ymax=0.65,
        width  = \columnwidth,
        height = 0.9*4.5cm,
        scaled y ticks = false,
        xlabel = {\# of batches},
        ymajorgrids = true,]
\addplot[sc6] table [x=Step, y=Value, col sep=comma] {figure-data/hard-labels/confidence_gap_with_aug/run-version_6_confidence_gap_addtocart-tag-confidence_gap.csv};
\addplot[sc7] table [x=Step, y=Value, col sep=comma] {figure-data/hard-labels/confidence_gap_with_aug/run-version_6_confidence_gap_cart-tag-confidence_gap.csv};
\addplot[sc8] table [x=Step, y=Value, col sep=comma] {figure-data/hard-labels/confidence_gap_with_aug/run-version_6_confidence_gap_mainpicture-tag-confidence_gap.csv};
\addplot[sc9] table [x=Step, y=Value, col sep=comma] {figure-data/hard-labels/confidence_gap_with_aug/run-version_6_confidence_gap_name-tag-confidence_gap.csv};
\addplot[sc10] table [x=Step, y=Value, col sep=comma] {figure-data/hard-labels/confidence_gap_with_aug/run-version_6_confidence_gap_price-tag-confidence_gap.csv};
\addplot[sc11] table [x=Step, y=Value, col sep=comma] {figure-data/hard-labels/confidence_gap_with_aug/run-version_6_confidence_gap_subjectnode_prod-tag-confidence_gap.csv};
\end{axis}
\end{tikzpicture}
\end{minipage}

\begin{tikzpicture}
\begin{axis}[
    hide axis,
    legend style={
        font = \small,
        legend columns=3, 
        at={(0.5,-0.3cm)}, 
        anchor=north,
    },
    xmin=0,
    xmax=1,
    ymin=0,
    ymax=1
]

\addplot[sc0] coordinates {(0,0)};
\addlegendentry{add-to-cart};
\addplot[sc1] coordinates {(0,0)};
\addlegendentry{go-to-cart};
\addplot[sc2] coordinates {(0,0)};
\addlegendentry{mainpicture};
\addplot[sc3] coordinates {(0,0)};
\addlegendentry{name};
\addplot[sc4] coordinates {(0,0)};
\addlegendentry{price};
\addplot[sc5] coordinates {(0,0)};
\addlegendentry{subject-node};

\end{axis}

\end{tikzpicture}

\caption{Effect on the average \emph{confidence gap} from using \emph{hard examples} (see section \ref{sec:hardlabels_tq}) for GCN-Mean. The left plot shows the average confidence gaps during training without augmentation. On the right, we show the corresponding confidence gaps when the hard examples are added to the training set.}
\label{fig:confidence_gap_no_aug}

\end{figure*}

\subsubsection{Effects on Nomination Accuracy}
In these experiments, we use the entire dataset with pages from all markets, with $M=20$, $K=5$, $T=50$ and $L=6$. Figure \ref{fig:aug_over_time} highlights the effect of the timing of the \emph{hard example} augmentation step on the average nomination accuracy on the validation set while training the FCN and GCN-Mean. We notice that the timing of the augmentation makes little difference on the final nomination accuracy. This could indicate that there are a few consistently confusing elements that can be relatively easily identified using models trained on random unlabelled elements as negatives. Once these elements are in the training set, the models' nomination performance abruptly improves. Indeed both FCN and GCN-Mean obtain lower errors after the augmentation, by about $2$ and $5$ p.p., respectively, GCN-Mean achieving beyond $80\%$ nomination accuracy and FCN exceeding $65\%$.

\subsubsection{Effects on the Quality of Rankings}\label{sec:hardlabels_tq}
Here we measure whether \TrainingName\ gets us closer to training directly on the nomination objective. As a proxy, we measure this as the difference between the confidence assigned to the labeled element and the highest confidence assigned to any unlabeled element on the same page. We call this difference the \emph{confidence gap}. We assume that the higher this difference is, the \emph{better} the resulting ranking would be. Since either unlabeled elements become less likely to be mistaken for the true element (when the true element is ranked first) or the difference by which an unlabeled element wins in the ranking (when the true element is not ranked first) is lower, and hence the ground-truth labeled element appears higher in the ranking. 

In Figure \ref{fig:confidence_gap_no_aug}, we ask the models to perform the nomination task on the entire training set every $50$ epochs and measure the average confidence gap per class. From the normal classification objective, we notice that the confidence gap is increasing on average, yet this behaviour is far from monotonic. In contrast, according to our metric, the rankings that the nomination objective is based on gets steadily better when we add \emph{hard examples} during training. Thus this performance resembles what we would expect when optimizing directly on the nomination objective. In this experiment, augmentation is performed every $50$ epochs, though the same behaviour is exhibited when this step is performed once. 

\section{Limitations}\label{sec:limitations}
One shortcoming of the \DatasetName\ is that far from all elements have a ground truth label. We only have one label per class per page, which makes the results slightly pessimistic. In reality, a click event on several elements in the \emph{buy button} (e.g. the button itself and the node containing the button text) could generate the same result. Thus, there are several acceptable nominations in practice. The \emph{single label per element type} issue reduces the number of positive training examples in the training set and increases the difficulty of the nomination task. While this results in a lower overall accuracy than what would be achieved in practice, it should not affect the comparison between methods. Furthermore, perfectly annotating all nodes of each element in a dataset of this size is very challenging, and would require substantial web-development knowledge. An approach for reducing the stringency of the evaluation would be to check the content of nominated elements. Still, this is not appropriate for action elements and requires resilient content extraction heuristics for all considered classes.

\section{Conclusion}\label{sec:conclusions}
We introduce the \DatasetName, a large-scale realistic dataset containing $51,701$ product pages from $8,175$ merchants across $8$ markets, with labels that present varied challenges. To initiate research on our dataset, we explored the potential of GNNs for nominating webpage elements and how these methods could be combined with LLMs. Our experiments underscore the untapped potential of the plain GCN-Mean algorithm, which we believe could serve as a component in more sophisticated nomination architectures.

\bibliography{mybib}
\bibliographystyle{abbrvnat}
\end{document}